\documentclass[numsec,webpdf,modern,medium]{oup-authoring-template}%

\usepackage{xcolor}

\usepackage{booktabs}
\usepackage{makecell} 
\usepackage{array} 

\usepackage{float}
\usepackage{caption}

\usepackage[T1]{fontenc}
\usepackage[utf8]{inputenc}
\usepackage{lmodern}
\usepackage{tikz}
\usetikzlibrary{arrows.meta,positioning,fit}

\graphicspath{{Fig/}}

\setcitestyle{round}
\begin{document}

\journaltitle{Authors' Original Manuscripts (version before submission)}
\DOI{DOI HERE}
\copyrightyear{November 2025}
\pubyear{2025}
\access{Advance Access Publication Date: Day Month Year}
\appnotes{Paper}

\firstpage{1}


\title[LLM poetry]{The author is dead, but what if they never lived? A reception experiment on Czech AI- and human-authored poetry}

\author[1]{Anna Marklová\ORCID{0000-0003-3392-1028}}
\author[2,$\ast$]{Ondřej Vinš\ORCID{0000-0002-4697-5351}}
\author[3]{Martina Vokáčová\ORCID{0000-0002-2396-2240}}
\author[1]{Jiří Milička\ORCID{}}

\authormark{Marklová, Vinš et al.}

\address[1]{\orgdiv{Department of Linguistics}, \orgname{Charles University}, \orgaddress{\street{nám. J. Palacha 1/2}, \postcode{11638}, \country{Czech Republic}}}

\address[2]{\orgdiv{Institute of Czech and Deaf studies}, \orgname{Charles University}, \orgaddress{\street{Na Příkopě 29}, \postcode{11000}, \country{Czech Republic}}}

\address[3]{\orgdiv{Department of Germanic and Nordic Studies}, \orgname{Charles University}, \orgaddress{\street{nám. J. Palacha 1/2}, \postcode{11638}, \country{Czech Republic}}}

\corresp[$\ast$]{Corresponding author. \href{email: ondrej.vins@ff.cuni.cz}{ondrej.vins@ff.cuni.cz}}

\received{Date}{0}{2025}
\revised{Date}{0}{Year}
\accepted{Date}{0}{Year}



\abstract{Large language models are increasingly capable of producing creative texts, yet most studies on AI-generated poetry focus on English --- a language that dominates training data. In this paper, we examine the perception of AI- and human-written Czech poetry. We ask if Czech native speakers are able to identify it and how they aesthetically judge it. Participants performed at chance level when guessing authorship (45.8\% correct on average), indicating that Czech AI-generated poems were largely indistinguishable from human-written ones. Aesthetic evaluations revealed a strong authorship bias: when participants believed a poem was AI-generated, they rated it as less favorably, even though AI poems were in fact rated equally or more favorably than human ones on average. The logistic regression model uncovered that the more the people liked a poem, the less probable was that they accurately assign the authorship. Familiarity with poetry or literary background had no effect on recognition accuracy. Our findings show that AI can convincingly produce poetry even in a morphologically complex, low-resource (with respect of the training data of AI models) Slavic language such as Czech. The results suggest that readers' beliefs about authorship and the aesthetic evaluation of the poem are interconnected.}


\maketitle

\section{Introduction}

Is authorship important to the enjoyment of art? This question is hardly new. In literary theory, the possibility of a work standing apart from its author has long been debated. A decisive impulse came from Roland Barthes in his famous essay, The Death of the Author (La mort de l'auteur, \cite{Barthes1968}). According to him, ``the image of literature to be found in contemporary culture is tyrannically centered on the author, his person, his history, his tastes" \citep[][p.16]{Barthes1968}. To end such tyranny, Barthes proposes a radical shift towards the reader. ``[H]e is only that someone who holds gathered into a single field all the paths of which the text is constituted"\citep[][p.16]{Barthes1968}, so that the author must die in the reader's interest.

In practice, however, the author seems hard to kill. The difficulty of separating a work from its maker is evident, for instance, from readers boycotting best-selling writers over offensive public stances or misconduct, or in seeking out books precisely because they are written by a favorite novelist or celebrity. These examples suggest that authorship shapes how audiences approach and value artworks.

Nowadays, as artificial intelligence enters the field of art, the question of authorship is primarily addressed with regard to human co-authors in the areas of ethics and law, while the centuries-long debate over authorship in literary theory seems to be overlooked. But if it holds that the author is dead, does it matter that they never lived? If the reader is the one to play the key role in constructing literary meaning, does the artificiality of a work's origin make any difference?

AI as an author is steadily improving, and growing research in the field of AI-generated texts has been bringing evidence that people struggle, or even fail, to distinguish AI-written texts from human-written ones \citep{PorterMachery2024}. This, at times, gives rise to fears of cultural ruin: the end of human creativity, the downfall of book publishing, and the purposelessness of writers \citep[e.g.,][]{Geller2025}.

These considerations lead us to two main questions: Is it really true that people cannot recognize AI-generated texts, even in highly creative genres? And does knowing who authored a text change how people experience it? We address these questions using poetry as a representative of a highly inventive, language-sensitive literary form. 

As recent studies have shown, artificial intelligence can generate poetry that is often indistinguishable from human-written poems, and in some cases, it is even preferred by native speakers \citep[e.g.,][]{PorterMachery2024, Kbis2020ArtificialIV}. However, these studies have been conducted almost exclusively in English, a language that holds a unique position due to its overwhelming dominance in the training data of LLMs. As a result, their findings cannot be readily generalized to other languages.  

It is therefore crucial to examine AI-generated poetry in languages that constitute less than 1\% of the typical training data \citep{johnson2022ghostmachineamericanaccent}, representing the vast majority of the world's languages. Our study addresses this gap by focusing on one such language: Czech. We investigate how proficient AI is at producing poetry in Czech and how its outputs are perceived by native speakers. Moreover, we test AI's performance across two distinct genres of Czech poetry: contemporary free verse, which generally lacks rhyme, and Czech nonsense poetry, characterized by playfulness, absurdity, and inventiveness. These qualities --- particularly creativity, humor, and linguistic flexibility --- are often considered uniquely human, and there is a question of whether LLMs are able to re/produce them.

In an online experiment, we presented poems authored by AI and by humans to Czech native speakers who were asked to guess the authorship. Then, they were asked to judge the poem on several evaluative scales. Our goal was to determine whether people can recognize AI-generated poems and evaluate them more or less favorably than human-written poems.

We argue that even if AI can generate texts indistinguishable from human writing, it need not spell the end of human creative endeavors, since the knowledge --- or even mere belief --- that a text is AI-generated shifts people's enjoyment.

\section{Studies on AI poetry}

In the wake of the latest developments in LLMs and their availability for public use, research into AI-generated poetry has expanded. However, these studies have predominantly focused on English. 

Early work, such as \citet{Schwartz2015}\citep[cf.][]{Linardaki2022}, utilized rule-based systems like RACTER and found that approximately 65\% of readers could not identify whether poems were human- or AI-written, although the study lacked experimental control. \citet{HopkinsKiela2017} employed an LSTM model and reported that human poems received slightly higher ratings for form, readability, and emotion, but that the most ``human-like'' and aesthetic poems were AI-generated; 48.6\% of human and 53.8\% of AI poems were misattributed. \citet{LauEtAl2018} tested Deep-speare with both crowdworkers (n = 1000) and an expert evaluator, finding near-chance discrimination accuracy (53\%) and mixed judgments: AI poems performed better on rhyme and meter, while human poems were rated higher on emotion and readability. Later work using GPT-2 \citep[e.g.,][]{Lc2021, KobisMossink2021, GunserEtAl2022} consistently revealed similar patterns. Discrimination accuracy ranged from about 50\% to 66\%, and human poems were generally rated higher in aesthetic and emotional dimensions. Comparisons between human-in-the-loop (with human intervention ) and human-on-the-loop (without human intervention ) approaches \citep{KobisMossink2021} showed only small differences in quality ratings and discrimination rates. Recently, \citet{PorterMachery2024} asked non-experts in poetry to guess the authorship of GPT-3.5-generated and human-written poems and discovered that their success rate was slightly below chance on average (46.6\%). Moreover, the participants judged the AI poems more favorably in terms of beauty and rhythm, which may have contributed to the incorrect assumption that they were human-written. 

\citet{HITSUWARI2023107502}, in one of the few studies of poetry in a language other than English, explored Japanese haiku. They measured participants' ability to recognize authorship and the beauty scores attributed to the haiku across three conditions: human-made, AI-generated without human intervention, and AI-generated with human intervention. They discovered that participants could not distinguish between human-made and AI-generated haiku. Moreover, the beauty rating of the AI-generated haiku with human intervention was the highest. \citet{GunserEtAl2022} conducted an experiment in German in which participants judged whether poetic continuations (18 in total) were written by GPT-2 or a human, rated their confidence, and evaluated stylistic quality. Participants showed low accuracy in distinguishing AI- from human-written texts but exhibited overconfidence in their judgments. AI-generated continuations were rated as less well-written, inspiring, fascinating, interesting, and aesthetic when compared to both human-written and original continuations.

In addition to the mentioned studies, research has also been dedicated to the stylistic and linguistic analysis of AI poetry.  Specifically, \citet{Xin2024EnglishPG} examined English AI poetry and concluded that while the generated poems are syntactically and rhythmically flawless, they lack emotional and thematic depth. \citet{Chen2024EvaluatingDI} evaluated the diversity of English AI poetry generated by different model types across structural, lexical, semantic, and stylistic dimensions, revealing that AI poetry exhibits less diversity than human poetry along several dimensions, concerning, for example, insufficient rhyming and semantic uniformity. \citet{Hamat2024} analyzed AI poems generated by GPT-4, GPT-3 and PaLM2 using six lexical matrices and found that, compared to human poems, AI-generated poetry is lexically inferior.

\section{Czech poetry}
Czech poetry encompasses a diverse range of movements and individual styles. For our experiment, we selected poems from two categories. First, we chose modern, contemporary poetry (called `modern' poetry in this paper), specifically, poems published in the magazine Psí víno over the last ten years. Psí víno is a digital curatorial platform for contemporary poetry. It focuses on original Czech and Slovak poetry as well as translations of international voices. Each thematic issue is prepared by a rotating team of curators, and each poem selection is accompanied by a curator's commentary that situates the poems in a broader cultural, social, or interdisciplinary context. We selected poems that were originally written in Czech, matched for length, and authored by both men and women. Some of them are critically acclaimed authors (such as Marie Iljašenko or Bernardeta Babáková), and some of them are emerging authors (Oliver Frolich). Their poems were mostly written in free verse.

Second, we chose nonsense Czech poetry for the experiment. Since this genre is a very specific part of literature, we will sum up the main features. Following a quote by the author of nonsense literature Anthony Burgess, nonsense `is a playful pragmatic way of interpreting the universe' \citep{Burgess1987}. When our cognitive perception proves insufficient to fully grasp everyday reality, nonsense and the feeling of absurdity emerge. As Cassie Westwood points out, `[n]onsense revivifies our perception of a world that exceeds our ability to apprehend it rationally'. \citep{Westwood2022}. Nonsense literature --- whose peculiar language (as seen, e.g., in Lewis Carroll's Alice's Adventures in Wonderland) might resemble gibberish --- is a vast category with unclear borders. For our experiment, we picked poems from the following three most prominent Czech nonsense poets:

Bohdan Vojtěch Šumavanský (1891--1969), a Prague based poet born as Theodor Adalbert Rosenfeld, is the author of Lomikel na dlásnech (1937, pen-name T. R. Field), a poetry book the main topics of which are macabre mystification and artificial mythology centered around a picturesque deity called Lomikel who resides in sewer lines. Field's poetry is typical of neologisms, which form a complex network within his writings but are unintelligible to regular speakers of Czech (cf. \citealp{Kovarik1994Field}). Emanuel Frynta (1923--1975) is the author of Závratné pomyšlení (1993, published posthumously) and Písničky bez muziky \citep{Frynta1988}. In his playful poems, he examines the possibilities and limits of language as a means of communication. Frequent inhabitants of these works are animals with the ability to speak (cf. \citealp[]{Zizler2006Frynta}). Josef Kainar (1917--1971) is the author of Nevídáno neslýcháno \citep{Kainar1964}, a poetry collection for children. In his poems for children, Kainar uses onomatopoeic expressions from nursery rhymes and anthropomorphised animals (cf.  \citealp{Opelik1994Kainar}).

\section{Methodology}
All materials, raw data, analysis scripts, and supplementary visualizations are publicly available at \href{https://osf.io/3awjt/overview?view_only=9444188b1bb64d0c918b5022548e7a04}{this link}. The study design was approved by the Ethics committee of the Faculty of Arts, Charles University.

\subsection{Aim and goals/research questions}
Building on the issues outlined in the Introduction, we formulated the following research questions:

RQ1: Can Czech native speakers recognize whether a poem is authored by a human or by AI? 

RQ2: Does the distinguishability of human- and AI-authored poems differ between modern and nonsensical Czech poetry (i.e., is one of these genres easier for AI to mimic convincingly)? 

RQ3: Do speakers' experience and engagement with poetry predict their ability to recognize poem authorship?

RQ4: Do Czech natives evaluate human- and AI-authored poems similarly in terms of overall liking and the attributes they assign to them?

\subsection{Materials}
The materials contained a total of 34 poems. Of these, 32 were critical items that were included in the analysis. The remaining 2 were control poems, after which a comprehension question followed. These poems were displayed midway through the experiment and at its conclusion. Since their purpose was to check participants'  attention, they were not included in the analysis.

\begin{figure*}[!htbp]
    \centering
    \includegraphics[width=1\textwidth]{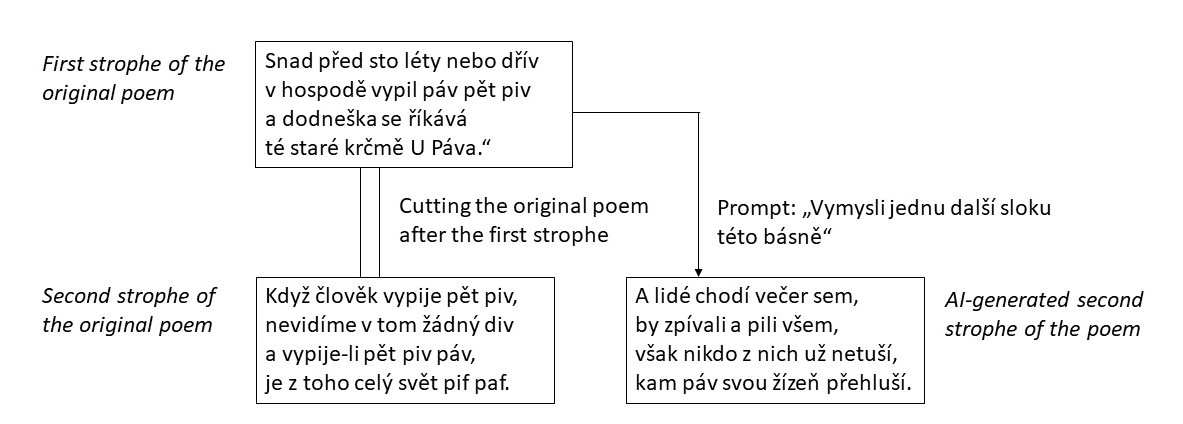}
    \caption{A scheme of creating the experiment materials.}
    \label{fig:poem_generating}
\end{figure*}

The 32 critical poems formed 16 logical pairs: each consisting of one human-written and one AI-generated poem. The poems were generated as follows (see \ref{fig:poem_generating} for better clarification). 

An existing Czech poem was selected and cut at a natural break, e.g., after a stanza. The first part of the poem was then supplied to GPT-4.5 Preview in chat mode together with a user prompt: ``Vymysli jednu další sloku této básně: [....]''  (Create another stanza for this poem). The system prompt was ``You are a proficient Czech poet.''  All AI continuations were generated in March 2025. 

The stanza generated by the model was used in the experiment without any modifications made by the experimenters. The stanza from the original poem that followed after the cut was also used as a test item; therefore, these two stanzas --- the human-written and the AI-generated continuation --- formed a logical pair. The experiment was designed in such way that one participant never saw both parts of one pair (see Procedure for details).

The rationale behind this method of poem generation was to explore what is most likely to end up in the so-called ``AI slop'' — that is, the kind of content expected to flood the internet in large quantities with minimal human effort. Therefore, our material does not represent the full creative potential of AI in poetry generation. It is reasonable to assume that with more time and effort invested in prompt engineering, poem selection, editing, and experimenting with different models and settings, the results could be substantially improved. However, for this initial experiment on Czech AI poetry, our goal was to first examine the kind of output that is expected to dominate online platforms.

To illustrate what the poems generated by GPT-4.5 Preview in chat mode looked like, we offer several examples. In the original Emanuel Frynta's poem \textit{Páv} `A Peacock', a regular rhyme scheme is to be observed: \textit{Když člověk vypije pět piv,/nevidíme v tom žádný div} `No wonder should feel his peers/when a man in a pub drinks five beers'\footnote{All poem examples were translated by Ondřej Vinš for the purpose of this article.}) (Frynta 1988). In the Chat-GPT's continuation, we see that the rhyme scheme is kept: \textit{A lidé chodí večer sem,/by zpívali a pili všem} `In the evening, people enter a beer hall/To sing and drink to all'. However, the semantics of the generated text might seem unusual to a native speaker of Czech (particularly the dative plural form of the pronoun \textit{všem} `all'). On the other hand,  nonsense poetry in general departs from standard language; therefore, such unusual word forms are not necessarily strange per se.

T. R. Field's poem \textit{Žalobej} (non-existing word in Czech) is an example of language nonsense. Almost the whole poem consists of pseudowords that resemble Slavic roots (e.g., \textit{duša} as an old word for \textit{duše} `soul'): \textit{da huhlavy mhlivy dvar zdaraváša dyjdi!} (non-existing Slavic-sounding words). Its tone is that of a sermon or prayer to god Lomikel (similar to the Lord's Prayer). It is worth noting that the GPT-4.5 generated continuation --- bearing no traits of a prayer --- contains real words in Slovak, the closest language to Czech: \textit{čolom ťažko znie môj bol, v srdci rana cenná!} (a mix of existing Czech, Slovak, and non-existing Slavic-sounding words).

The title of Josef Kainar's poem \textit{Abraka dabraka} refers to the ancient magic word \textit{abracadabra}, of (probably) Aramaic origin. Kainar's usage alludes to nursery rhymes that are common among Czech children. His language is highly onomatopoeic, employing internal rhyme at the level of a line: \textit{mudr a judr/z říše plyše} `Mudr. and Judr./from the land of plush'\citep{Kainar1964}. This specific feature can also be found in the GPT-4.5 continuation, even in a more prominent way: \textit{hopsa hejsa, ťapi ťapi, šušky mušky} (Czech onomatopoeia).

In the modern poem \textit{Muzeum medúz} `The Jellyfish Museum', by Czech-Ukrainian poet Maria Iljašenko, the images of luminescent jellyfish floating in tanks are the scene where two old acquaintances meet after a long time. Dead bodies of jellyfish may allude to the war memorial, located in front of the Museum, thus emphasizing the anti-war meaning of the poem. Although the original poem is in free verse, the AI-generated continuation is rhymed, stressing the brief moment of an encounter but using a basic rhyme pattern that resembles clichéd romantic poetry (the time stopped when I looked into your eyes, etc).

\subsection{Participants}
The participants were recruited from a pool of university students and from the authors' circle of friends and acquaintances. In total, 132 people participated in the study; however, 6 of them were excluded because they failed to answer one or more control comprehension questions correctly. The final, further analyzed sample thus consisted of 126 participants.

\begin{table*}[htbp]
\centering
\caption{Demographic characteristics of participants in Experiment~2 ($N = 126$).}
\begin{tabular}{ll}
\toprule
\textbf{Characteristic} & \textbf{Distribution} \\
\midrule
Gender & Female: 96 (76.2\%), Male: 28 (22.2\%), Non-binary: 2 (1.6\%) \\
Education & Secondary: 75 (59.5\%), University: 51 (40.5\%) \\
Age & $M = 25.87$, $SD = 7.66$, Range = 20--63 \\
Reading frequency & Never (0): 18 (14.3\%), Rarely (1): 60 (47.6\%), \\
& Sometimes (2): 36 (28.6\%), Often (3): 11 (8.7\%), Very often (4): 1 (0.8\%) \\
Poetry appreciation & Low (0--1): 15 (11.9\%), Medium (2): 62 (49.2\%), High (3): 49 (38.9\%) \\
Literary background & None: 32 (25.4\%), Basic: 74 (58.7\%), Intermediate: 11 (8.7\%), \\
& Advanced: 1 (0.8\%), Professional: 8 (6.3\%) \\
\bottomrule
\end{tabular}
\end{table*}

A total of 126 participants (76\% female; 22\% male; 2\% non-binary, $M_{age}$ = 25.9 years, range 20--63) took part in the study. For most participants, the highest level of education completed was high school (60\%), 40\% finished university education. The frequency of reading poetry was generally low, with nearly half of the respondents reporting that they rarely read poetry. Poetry appreciation was moderate to high, with 88\% scoring in the medium or high range. A majority (59\%) reported having some basic background in poetry, while only a small fraction had an advanced or professional background.

\subsection{Procedure}
The study was conducted online using a dedicated web interface. Participants received a link to the experiment, which they could complete at home or in another quiet environment. The entire procedure, including instructions and the demographic questionnaire, was presented in Czech. In the initial instructions, participants were asked to complete the experiment on a computer rather than on a smartphone or tablet to ensure, as far as possible, consistent testing conditions.

At the beginning of the experiment, participants viewed a welcome screen that displayed the study title, a brief description, and basic instructions. They were then provided with a link to a PDF version of the informed consent form and were asked to confirm that they had read it and agreed to participate by checking the box ``I have read the Informed Consent and I agree'' before proceeding further.

This was followed by a demographic questionnaire in which participants provided information on their gender, age, and education level. After filling out these basic demographic details, participants provided information about their experiences and attitudes towards poetry. Specifically, they were asked: (1) How often do you read or engage with poetry? (Never, Rarely, Sometimes, Often, Very often); (2) How much do you value poetry? (I don't care at all, A little, Quite a bit, Very much); (3) What education in the field of poetry have you received? (None or only from standard school classes, Self-study [hobby, passion], Bachelor's level [i.e., poetry was part of the curriculum], Master's level or higher); and (4) What is your background in poetry? (No background, Occasional reader, Amateur poet, Published poet, Professional—e.g., teacher of language or literature, scholar, or editor).

In the main part of the experiment, participants were presented with a piece of a poem displayed in the center of the screen. Below each text, they were asked to indicate who they thought had written the poem (a human vs. artificial intelligence) and how confident they were in their answer, using a five-point scale ranging from very uncertain to very certain. Participants then evaluated each poem on several dimensions using five-point Likert-type scales. 
Specifically, they rated whether they liked the poem, whether it rhymed, and whether they perceived it as playful, imaginative, meaningful, and serious. All these evaluations were made on a scale from No --- Rather no --- I can't decide --- Rather yes --- Yes. After submitting their responses, the next poem appeared automatically.

The four evaluative attributes used to rate the poems were selected for the following reasons:
(1) \textit{hravá} `playful' is an adjective explicitly associated with nonsense poetry. It is also strongly linked to poetry for children, which overlaps to a certain degree with nonsense poetry, particularly in the works of two of our selected authors, Frynta and Kainar.
(2)	\textit{nápaditá} `imaginative' was included as a more general feature of poetry as a specific use of language (the idea that goes back to Russian formalism and Viktor Shklovsky's term \textit{ostranenie} `estrangement'), but it is also assigned as an important feature of nonsense poetry specifically.
(3)	\textit{smysluplná} `making sense' and (4) \textit{vážná} 'serious', on the other hand, are both adjectives more likely to be attributed to non-nonsense, modern poetry, at least in our preliminary expectations. We are aware that `sense' is a difficult concept to capture and that nonsense poetry often pushes the limits of sense-making in ways that readers may still experience as meaningful. Nonetheless, including this attribute allowed us to see how participants applied it when evaluating both modern and nonsense poems. 

The question about rhyme was included because the human-authored nonsense poems we used as prompts rhymed more often than the modern ones; therefore, we were interested in whether the model could rhyme in Czech successfully.

Each participant evaluated 16 critical poems. After 8 poems, a control item appeared: a poem similar in style to the critical ones, but after the usual ratings, a comprehension question appeared on the screen asking what the poem was about. A second control poem, accompanied by a comprehension question, was presented at the end of the experiment. 
The participants who failed to answer correctly to both of those questions were excluded from the analysis.

At the end of the experiment, participants were asked whether they recognized any of the poems (yes --- not sure --- no) and whether they were familiar with the work of T. R. Field, Josef Kainar, and Emanuel Frynta. We did not ask about the modern poets since they are not a part of the elementary and high school literature curriculum. After answering these questions, the responses were sent to the server, and the participants were shown a comprehensive overview of their responses, along with the percentage of their success.

\section{Results and discussion}

\subsection{Mixed effects logistic regression}

\begin{figure}[htbp]
    \includegraphics[width=0.48\textwidth]{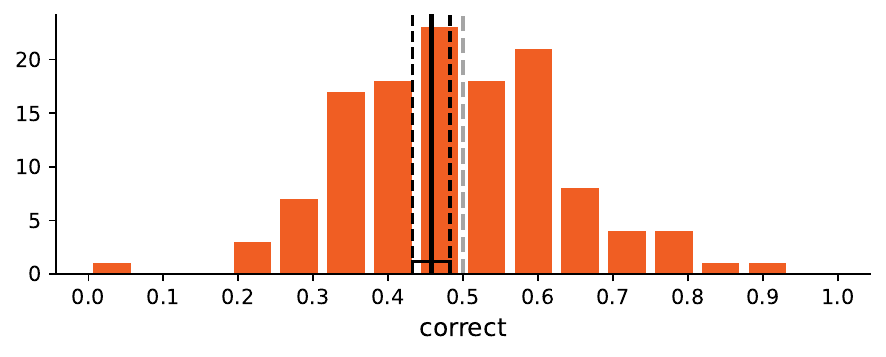}
       \caption{Distribution of correctness rates across participants. The x-axis represents the proportion of correctly identified poems, and the y-axis shows the number of participants.}
    \label{fig:correct_overall}
\end{figure}

Our first research question (RQ1) asked whether Czech native speakers can correctly recognize poem authorship. Using bootstrapped confidence intervals, we found that the overall correctness rate was on average 45.8\% (95\% CI: 43.3\%--48.3\%), which is slightly below the chance level of 50\%. This suggests that participants, on the whole, were not able to reliably distinguish between human- and AI-authored poems. The distribution of individual responses is shown in Figure~\ref{fig:correct_overall}. When we examined the results by poem type, we found that participants performed slightly above chance on average in the Nonsense category (51.4\%, 95\% CI: 48.3\%--54.5\%), whereas in the Modern category they were markedly less accurate (40.2\%, 95\% CI: 36.7\%--43.8\%).

Bootstrapped confidence intervals allowed us to estimate the overall correctness rate. To examine which factors influenced participants' guesses, we employed a mixed-effects logistic regression model with participant-level random effects. 
Fixed effects comprised participants' authorship judgments (human vs. AI), their confidence in those judgments, and, furthermore, poem ratings on several attributes (overall liking, rhyme, playfulness, imaginativeness, making sense, and seriousness). Demographic variables and information about poetry background were included as covariates. The dataset comprised 2,016 observations from 126 Czech participants. The full results are presented in Table \ref{tab:new-model}, with statistically significant effects marked in bold.

\begin{table*}[ht!]
\centering
\caption{
{\bf Mixed effects logistic regression results with estimates, confidence intervals, exponentiated coefficients, and significance levels.}
}
\label{tab:new-model}
\begin{tabular}{ll l l l l l l l l}
\hline
&&&\multicolumn{2}{l}{\bf Conf. Interval} & &\multicolumn{2}{l}{\bf Conf. Interval} &&\\
{\bf Variable} &{\bf Estimate} &{\bf SE} &{\bf Lower} &{\bf Upper} &{\bf Exp(B)} &{\bf Lower} &{\bf Upper} &{\bf z} &{\bf p }\\ 
\hline
(Intercept) & -0.15004 & 0.16362 & -0.47073 & 0.17066 & 0.861 & 0.625 & 1.186 & -0.9170 & 0.359\\
Nonsense vs. Modern & 0.25326 & 0.12650 & 0.00533 & 0.50119 & 1.288 & 1.005 & 1.651 & 2.0021 & \textbf{0.045}\\
Human authorship & 0.27241 & 0.11595 & 0.04516 & 0.49966 & 1.313 & 1.046 & 1.648 & 2.3494 & \textbf{0.019}\\
Confidence & -0.07744 & 0.06844 & -0.21157 & 0.05670 & 0.925 & 0.809 & 1.058 & -1.1315 & 0.258\\
Liking & -0.13105 & 0.06542 & -0.25926 & -0.00284 & 0.877 & 0.772 & 0.997 & -2.0033 & \textbf{0.045}\\
Rhyme & 0.02459 & 0.07546 & -0.12331 & 0.17250 & 1.025 & 0.884 & 1.188 & 0.3259 & 0.744\\
Playful & 0.12822 & 0.05850 & 0.01355 & 0.24288 & 1.137 & 1.014 & 1.275 & 2.1915 & \textbf{0.028}\\
Imaginative & 0.07235 & 0.06845 & -0.06181 & 0.20652 & 1.075 & 0.940 & 1.229 & 1.0569 & 0.291\\
Making sense & 0.02451 & 0.05580 & -0.08487 & 0.13388 & 1.025 & 0.919 & 1.143 & 0.4391 & 0.661\\
Serious & -0.13825 & 0.05534 & -0.24671 & -0.02980 & 0.871 & 0.781 & 0.971 & -2.4985 & \textbf{0.012}\\
Trial number & -0.00923 & 0.01156 & -0.03189 & 0.01344 & 0.991 & 0.969 & 1.014 & -0.7980 & 0.425\\
Answer order & -0.04236 & 0.12646 & -0.29022 & 0.20551 & 0.959 & 0.748 & 1.228 & -0.3349 & 0.738\\
Poetry frequency & 0.11908 & 0.10116 & -0.07919 & 0.31736 & 1.126 & 0.924 & 1.373 & 1.1772 & 0.239\\
Poetry appreciation & 0.01758 & 0.10401 & -0.18628 & 0.22144 & 1.018 & 0.830 & 1.248 & 0.1690 & 0.866\\
Poetry background & 0.02882 & 0.08399 & -0.13580 & 0.19344 & 1.029 & 0.873 & 1.213 & 0.3431 & 0.731\\
Poet familiarity & -0.03708 & 0.13328 & -0.29831 & 0.22414 & 0.964 & 0.742 & 1.251 & -0.2782 & 0.781\\
Poem familiarity & 0.09216 & 0.14077 & -0.18375 & 0.36807 & 1.097 & 0.832 & 1.445 & 0.6546 & 0.513\\
Gender male --- female & 0.12096 & 0.14085 & -0.15511 & 0.39702 & 1.129 & 0.856 & 1.487 & 0.8588 & 0.390\\
Gender non-binary --- female & -0.00765 & 0.46076 & -0.91072 & 0.89543 & 0.992 & 0.402 & 2.448 & -0.0166 & 0.987\\
University --- secondary ed. & 0.11777 & 0.13129 & -0.13955 & 0.37509 & 1.125 & 0.870 & 1.455 & 0.8970 & 0.370\\
Age & 0.00557 & 0.00848 & -0.01105 & 0.02218 & 1.006 & 0.989 & 1.022 & 0.6565 & 0.511\\
\hline
\end{tabular}
\end{table*}

In our second research question (RQ2), we investigated whether AI is capable of generating modern as well as nonsensical poetry in Czech that is indistinguishable from that written by humans, and whether one of these genres is easier to generate convincingly. The model revealed that poem category influenced responses: nonsense poems were somewhat easier to recognize as AI-generated than modern poems ($\chi^{2}(1) = 4.01$, $p = .045$). Therefore, we can conclude that it was easier for LLMs to generate convincing modern Czech poems than nonsense ones. We will describe the details of the attributes given to the poems later in this section.

In addition to the poem category, the analysis uncovered that the fixed effect of actual human authorship was significant ($\chi^{2}(1) = 5.52$, $p = .019$), with human-written poems being recognized more accurately than AI-generated ones (Estimate = 0.27, 95\% CI [0.046, 0.500], OR = 1.31). This suggests that when people are confronted with a human poem, they are more likely to correctly assign it to its human author; however, when it is authored by AI, they struggle more to recognize it. 

Beyond the poem category and authorship, several of the evaluative dimensions also significantly influenced recognition accuracy. We will now comment on these predictors in turn: 
\begin{itemize}
    \item \textbf{Liking} negatively predicted correct classification ($\chi^{2}(1) = 4.01$, $p = .045$). Poems that participants liked more were paradoxically less likely to be correctly attributed.
    
    \item \textbf{Playfulness} positively predicted correct classification ($\chi^{2}(1) = 4.80$, $p = .028$). The more playful a poem was perceived to be, the more likely participants were to identify its true authorship.
    
    \item \textbf{Seriousness} negatively predicted classification ($\chi^{2}(1) = 6.24$, $p = .012$). Poems judged to be more serious were classified with less accuracy.
\end{itemize}

These findings are consistent with our earlier observation that nonsense poetry was easier to classify correctly than modern poetry. Nonsense poetry is characteristically perceived as more playful and less serious. It is plausible, then, to suppose that these two attributes may have served as clues in participants' judgments.  
To investigate this further,  we again employed bootstrapped confidence interval analysis with visualization (see the following section).

None of the remaining variables reached statistical significance (Table 2). Demographic variables (gender, age, education) showed no evidence of an association with classification accuracy. Moreover, measures of poetry engagement --- prior involvement with poetry (poetry background), self-reported frequency of reading poetry (poetry frequency), and general liking and interest in poetry (poetry appreciation) --- likewise did not have an effect. We also found no evidence that familiarity with the poems or with the poets used in the experiment (Kainar, Field, Frynta) improved performance.

These findings answer our third research question (RQ3), confirming that participants' or familiarity with poetry do not affect the correctness of guessing whether a poem is written by AI or a human.

We also controlled for design-related variables: trial number and answer order. In the experiment, the order of poems was shuffled for each participant, and the left-right mapping of response options (author human vs. AI) was counterbalanced between participants 
We discovered that none of these features had an effect on correctness.

Lastly, self-reported confidence did not correlate with the correctness of guesses. When people stated that they were confident in their choice, they had a similar probability of being correct as when they answered that they were not confident.


\subsection{Assessment of poetry}
In our fourth research question (RQ4), we asked whether Czech native speakers evaluate human poetry similarly to AI poetry, in terms of how much they like it and what attributes they assign to it. We used bootstrapped confidence intervals and examined the assessments of various features.

\subsubsection{Do people like human poetry more than AI poetry?}

\begin{figure}[htbp]
    \includegraphics[width=0.48\textwidth]{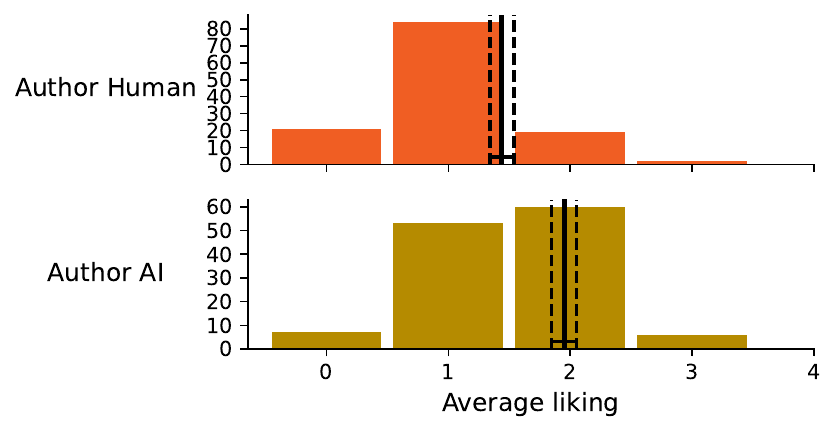}
       \caption{Average liking according to poem authorship}
    \label{fig:liking_isauthorhuman}
\end{figure}

When we look at the ridgelines showing the average liking of poems, separated by authorship and category of the poems (Figure~\ref{fig:liking_isauthorhuman}), we see that poems authored by AI were rated higher on average ($M = 2.0$, 95\% CI [1.8, 2.1]) than poems by humans ($M = 1.4$, 95\% CI [1.3, 1.5]).

\begin{figure}[htbp]
    \includegraphics[width=0.48\textwidth]{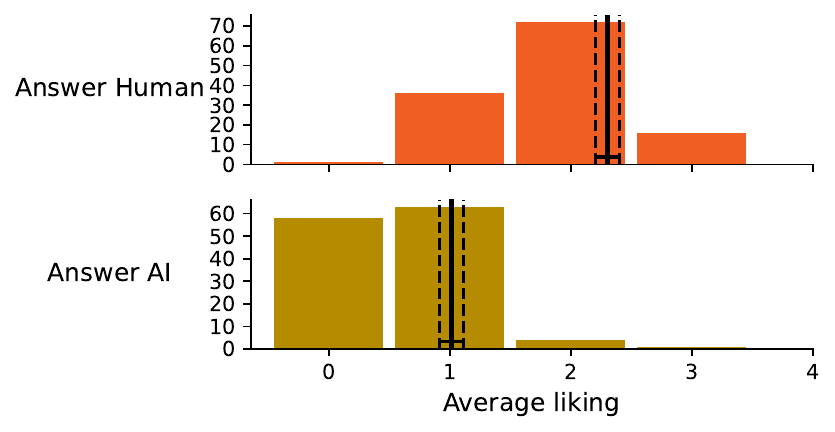}
       \caption{Average liking according to perceived poem authorship.}
    \label{fig:liking_isanswerhuman}
\end{figure}

However, when we look at how people answered, we see that when participants thought the poem was authored by a human, they liked it more ($M = 2.3$, 95\% CI [2.2, 2.4]) than when they thought it was authored by AI ($M = 1.0$, 95\% CI [0.9, 1.1]; Figure~\ref{fig:liking_isanswerhuman}). Since we do not know the direction of this correlation, it could also be the other way around: when people liked the poem, they may have been more inclined to think it was authored by a human rather than by AI. This discrepancy is reflected in the fact that, in the mixed model, ``liking" was found to have negatively influenced the chance of correct guesses.

\begin{figure}[htbp]
    \includegraphics[width=0.48\textwidth]{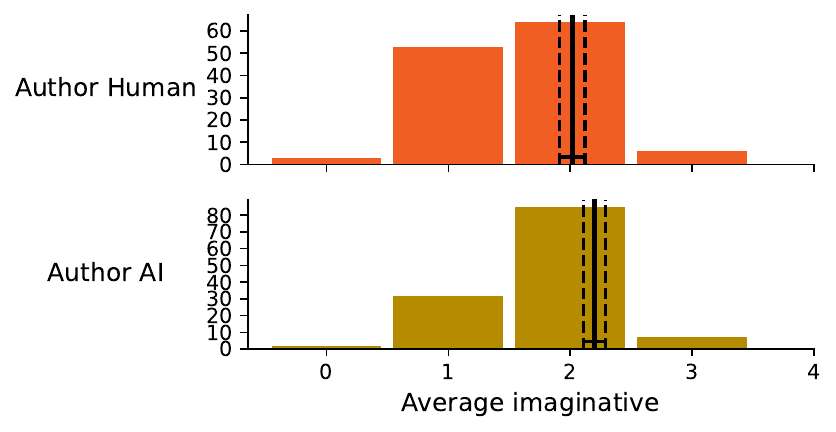}
       \caption{Average imaginativeness according to poem authorship}
    \label{fig:imaginative_isauthorhuman}
\end{figure}

\begin{figure}[htbp]
    \includegraphics[width=0.48\textwidth]{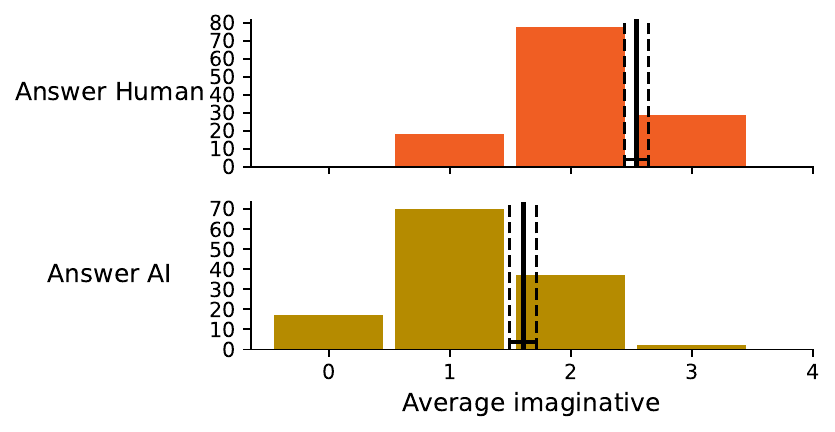}
       \caption{Average imaginativeness according to perceived poem authorship}
    \label{fig:imaginative_isanswerhuman}
\end{figure}

We observed a similar tendency in other rating scales. For the rating of how imaginative (\textit{nápadité}) the poems are, the actual authorship did not show a notable difference (Figure~\ref{fig:imaginative_isauthorhuman}): poems authored by AI were rated only slightly higher ($M = 2.2$, 95\% CI [2.1, 2.3]) than poems authored by humans ($M = 2.0$, 95\% CI [1.9, 2.1]). However, when participants thought that a poem was authored by AI, they rated it as significantly less imaginative ($M = 1.6$, 95\% CI [1.5, 1.7]) than when they thought it was authored by a human ($M = 2.5$, 95\% CI [2.4, 2.6]; Figure~\ref{fig:imaginative_isanswerhuman}).


\begin{figure}[htbp]
    \includegraphics[width=0.48\textwidth]{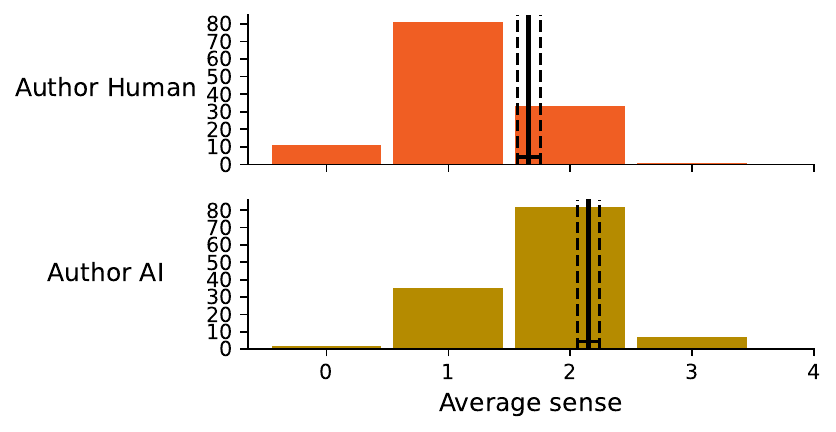}
       \caption{Average making sense according to poem authorship}
    \label{fig:sense_isauthorhuman}
\end{figure}

\begin{figure}[htbp]
    \includegraphics[width=0.48\textwidth]{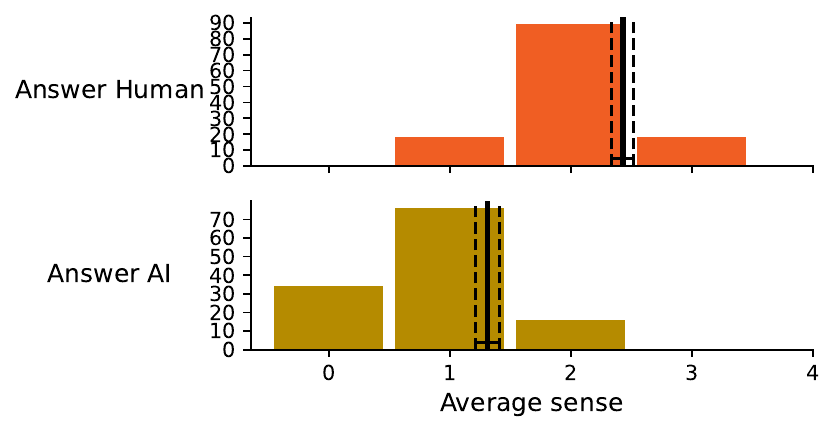}
       \caption{Average making sense according to perceived poem authorship}
    \label{fig:sense_isanswerhuman}
\end{figure}

Similarly, when we asked how much the poem makes sense (\textit{smyslusplná}), the actual authorship showed the following significant difference (Figure~\ref{fig:sense_isauthorhuman}): poems authored by AI were rated as more sensible ($M = 2.1$, 95\% CI [2.1, 2.2]) than those authored by humans ($M = 1.7$, 95\% CI [1.6, 1.8]). However, when participants believed that a poem was authored by a human, they rated it as making significantly more sense ($M = 2.4$, 95\% CI [2.3, 2.5]) than when they thought it was authored by AI ($M = 1.3$, 95\% CI [1.2, 1.4]; Figure~\ref{fig:sense_isanswerhuman}).

We found the analogous pattern with the other scales as well, the figures and CIs can be found in Supplementary materials. These findings indicate that bias regarding authorship, rather than authorship itself, drives evaluation. Participants penalize texts if they think they are AI-generated. However, when we look at the actual authorship, the AI-generated poems are evaluated similarly to those written by humans, or AI tends to score higher. 

We emphasize that the direction of this correlation cannot be determined from our data: people either think about a poem that it is authored by AI and, therefore, like it less, or they do not like a poem very much and, therefore, think it is authored by AI. Both processes may operate simultaneously, and they may differ across individuals.

In the next section, we will focus on the scales in more detail with respect to the differences between modern and nonsense poetry.

\subsection{Can AI mimic nonsense features?}
To assess whether AI can mimic nonsense poetry and its specific features, we employed bootstrapped confidence intervals. In this analysis, 
we compared rating profiles across four conditions: modern poems written by AI, modern poems written by humans, nonsense poems written by AI, and nonsense poems written by humans. By that, we can compare the ratings of AI nonsense poetry with both nonsense human poetry (expected to be rated similarly) and modern AI poetry (expected to be rated lower on the scales indicating nonsense poetry). Additionally, it is important to determine whether human-written nonsense poetry is perceived differently from modern human-written poetry, therefore, whether the theoretical attributes of nonsense poetry are even perceived by people.

In the rating scales, we used these attributes: nonsense poetry is described as playful (\textit{hravá}) and imaginative (\textit{nápaditá}), therefore, these categories should be rated higher on the scales for nonsense poetry compared to modern. On the other hand, nonsense poetry should not make sense (\textit{smysluplná}) and is not serious (\textit{vážná}), therefore, we expect these scales to be rated lower for nonsense poetry compared to modern poetry.

Additionally, we added a scale for rating the rhyming, since the selected nonsense poems rhymed more often than the modern ones. We wanted to see if AI is able to rhyme in Czech.



\begin{figure}[htbp]
    \includegraphics[width=0.48\textwidth]{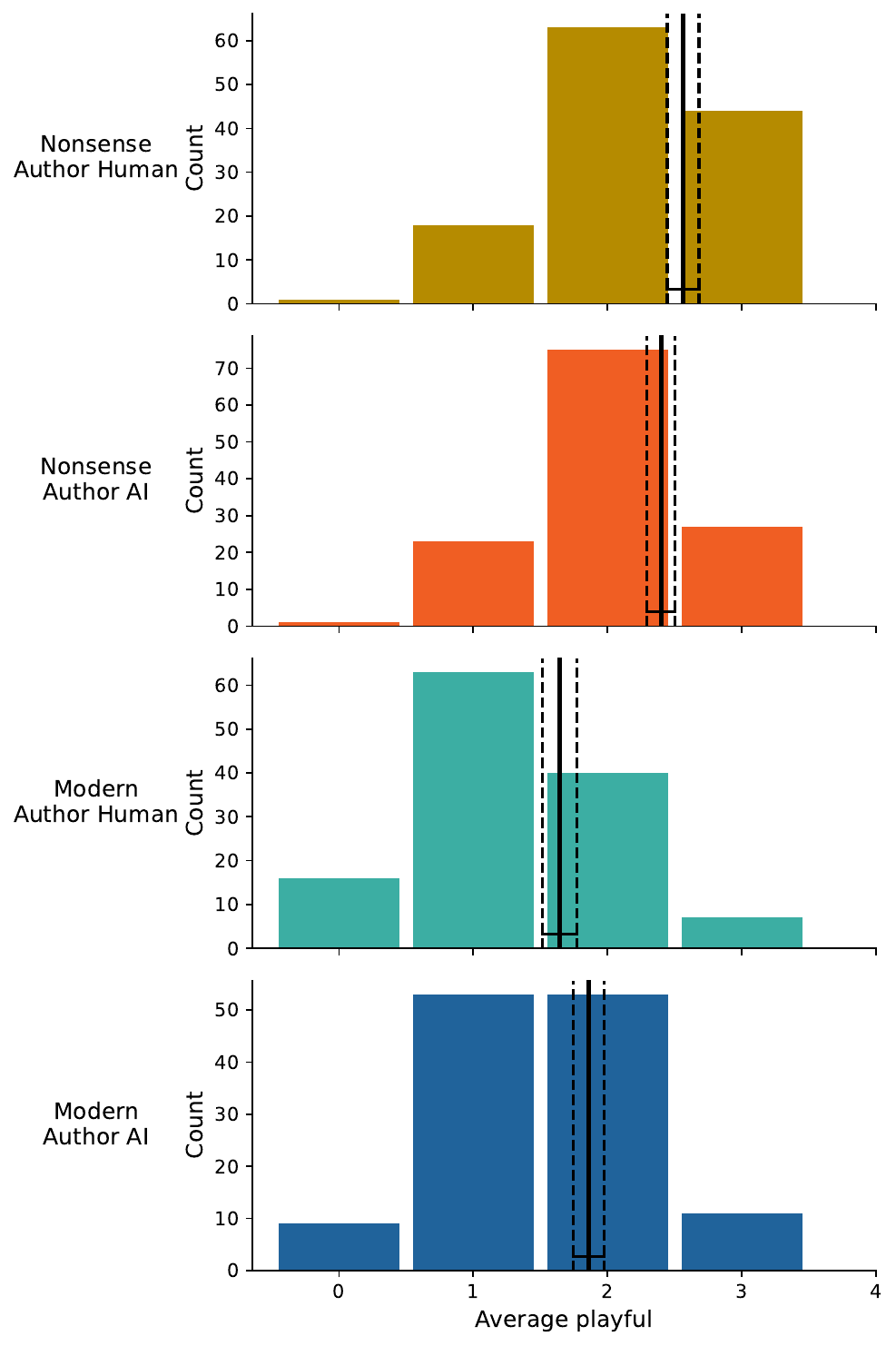}
       \caption{Average playful according to poem authorship}
    \label{fig:playful_nonsense}
\end{figure}

\begin{figure}[htbp]
    \includegraphics[width=0.48\textwidth]{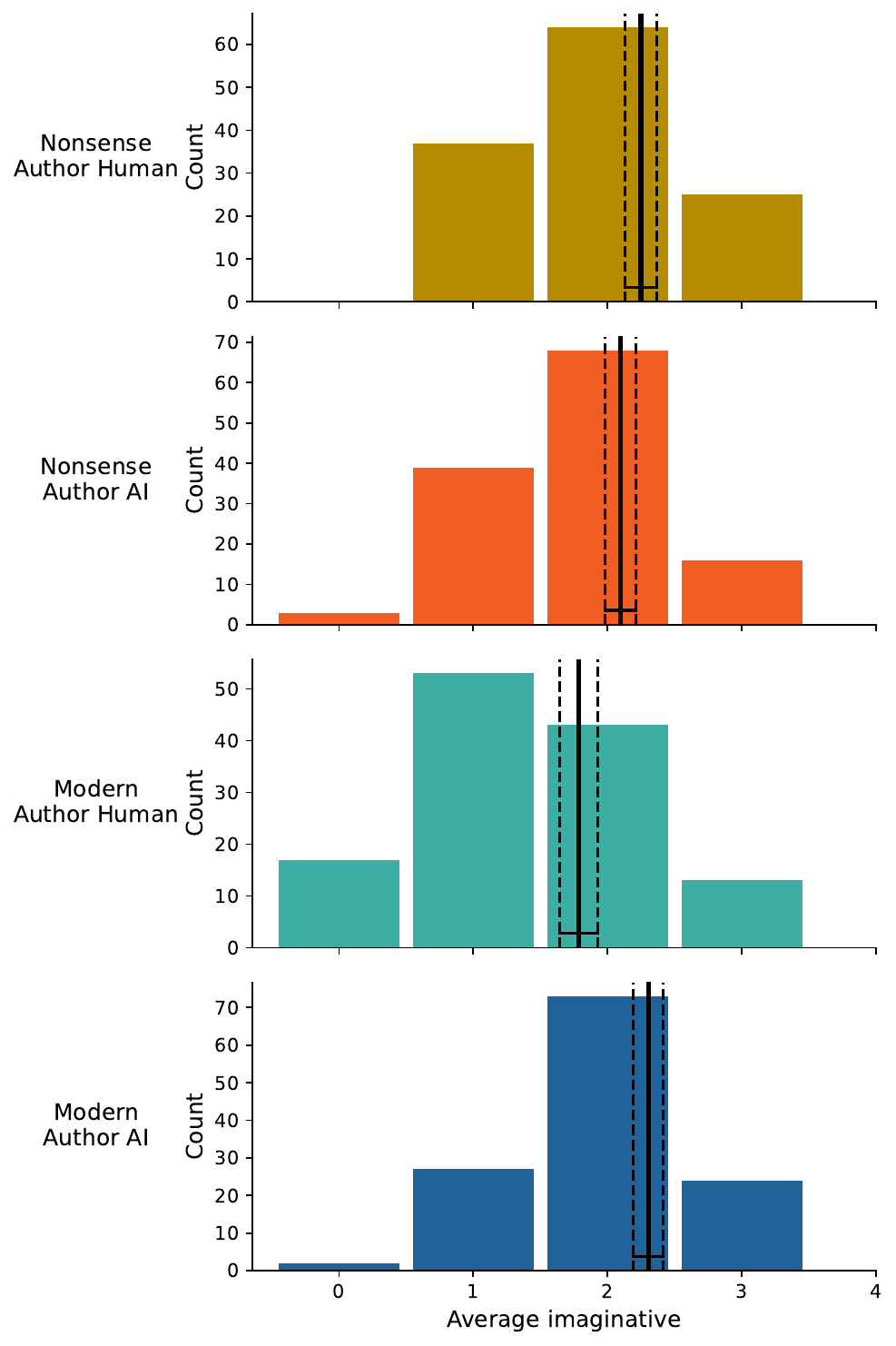}
       \caption{Average imaginative according to poem authorship}
    \label{fig:imaginative_nonsense}
\end{figure}

Firstly, we examined the attributes that we expected to be rated higher for nonsense poems. Figure~\ref{fig:playful_nonsense} shows ratings of playfulness, and Figure~\ref{fig:imaginative_nonsense} shows ratings of imaginativeness. We can see that AI successfully imitated the playfulness of nonsense poems and scored comparably to human-authored nonsense poems ($M = 2.40$, 95\% CI [2.29, 2.50] for AI; $M = 2.56$, 95\% CI [2.44, 2.68] for human). Modern poems authored by both humans and AI were rated considerably lower on this scale ($M = 1.86$, 95\% CI [1.74, 1.98] for AI; $M = 1.64$, 95\% CI [1.51, 1.77] for human).  

When we look at imaginativeness, we see that there is not much of a difference between modern and nonsense poems overall (Figure~\ref{fig:imaginative_nonsense}). Nonsense poems authored by humans ($M = 2.25$, 95\% CI [2.13, 2.37]) and by AI ($M = 2.10$, 95\% CI [1.98, 2.21]) were rated similarly imaginative, suggesting that AI was able to mimic this aspect of the original nonsense poems. In modern poetry, it appears that AI also implemented this aspect effectively and was rated as more imaginative ($M = 2.30$, 95\% CI [2.19, 2.41]) than modern human poetry ($M = 1.78$, 95\% CI [1.65, 1.93]).

\begin{figure}[htbp]
    \includegraphics[width=0.48\textwidth]{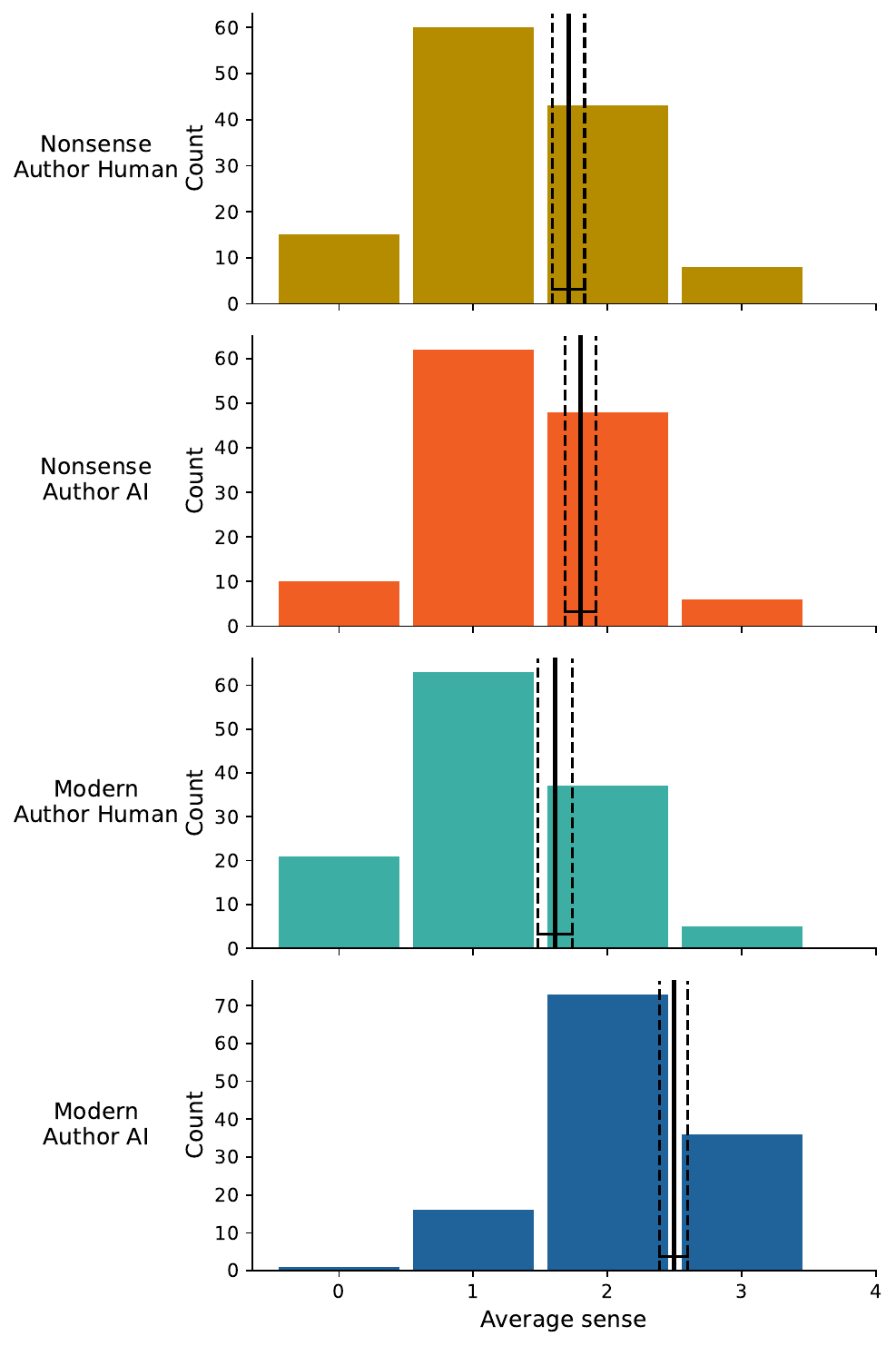}
       \caption{Average making sense according to poem authorship}
    \label{fig:sense_isauthorhuman_category}
\end{figure}

We expected that nonsense poetry will be rated lower for making sense. As shown in Figure~\ref{fig:sense_isauthorhuman_category}, nonsense AI poems ($M = 1.80$, 95\% CI [1.69, 1.92]) were indeed rated significantly lower than modern AI poems ($M = 2.49$, 95\% CI [2.39, 2.60]). Interestingly, there is no significant difference between modern human and nonsense human poems; both are rated comparably low in terms of making sense compared to nonsense AI poems (nonsense human $M = 1.71$, 95\% CI [1.59, 1.83]; modern human $M = 1.61$, 95\% CI [1.48, 1.74]).

\begin{figure}[htbp]
    \includegraphics[width=0.48\textwidth]{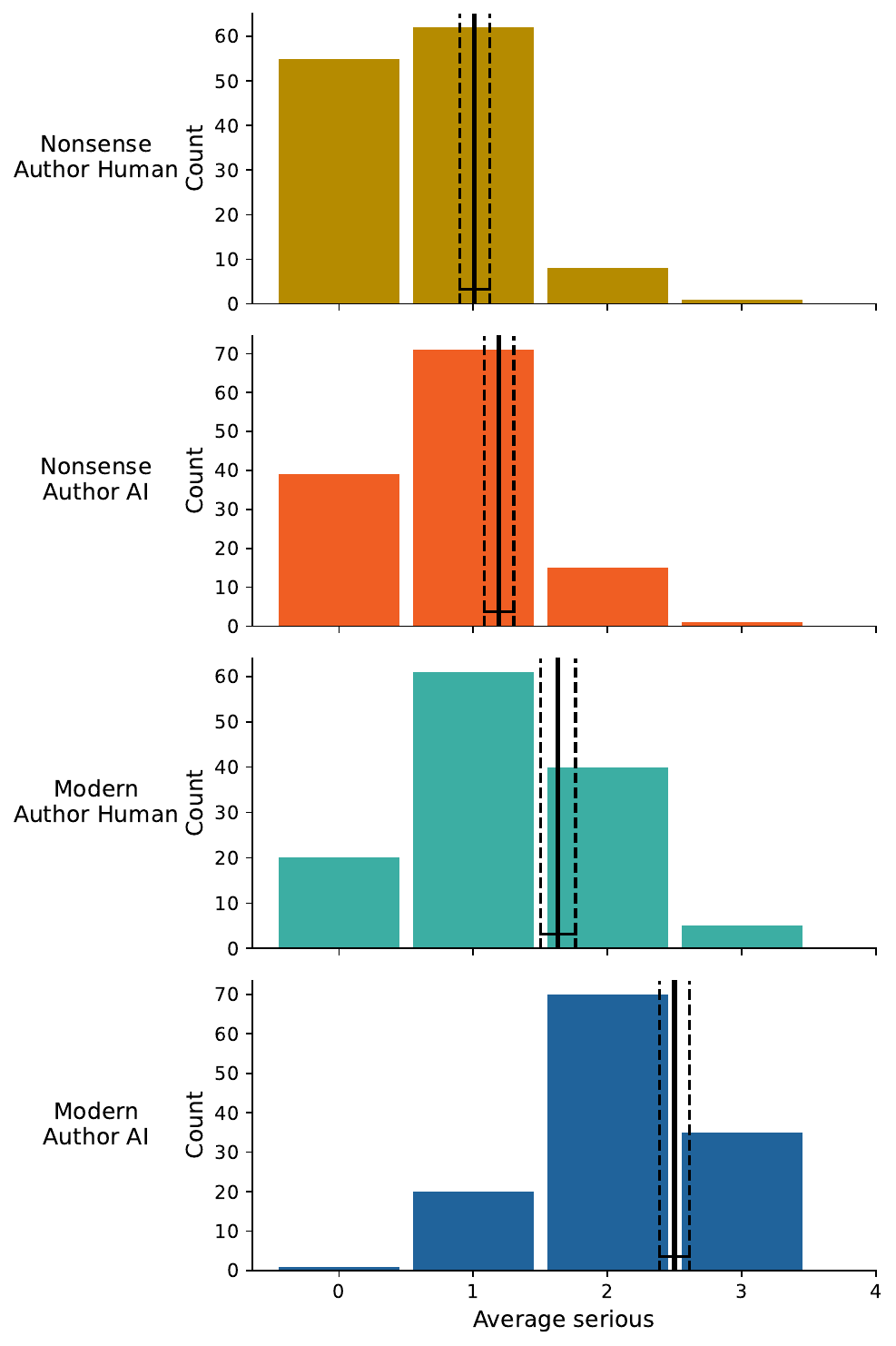}
       \caption{Average serious according to poem authorship}
    \label{fig:serious_isauthorhuman_category}
\end{figure}

For seriousness (Figure~\ref{fig:serious_isauthorhuman_category}), nonsense poems authored by both AI and humans were rated as less serious than modern poems, with human nonsense poems scoring the lowest (human nonsense $M = 1.01$, 95\% CI [0.90, 1.12]; AI nonsense $M = 1.19$, 95\% CI [1.08, 1.30]). Modern poems, on the other hand, were perceived as substantially more serious, particularly those generated by AI ($M = 2.50$, 95\% CI [2.39, 2.61]) compared to modern human poems ($M = 1.63$, 95\% CI [1.50, 1.76]).

Taken together, our findings suggest that 1) even human-written nonsense poetry is not consistently perceived as fulfilling its attributes compared to modern poetry, and 2) AI consistently mimics the attributes of nonsense poetry and often even exaggerates these attributes in comparison to human nonsense poetry.

\begin{figure}[htbp]
    \includegraphics[width=0.48\textwidth]{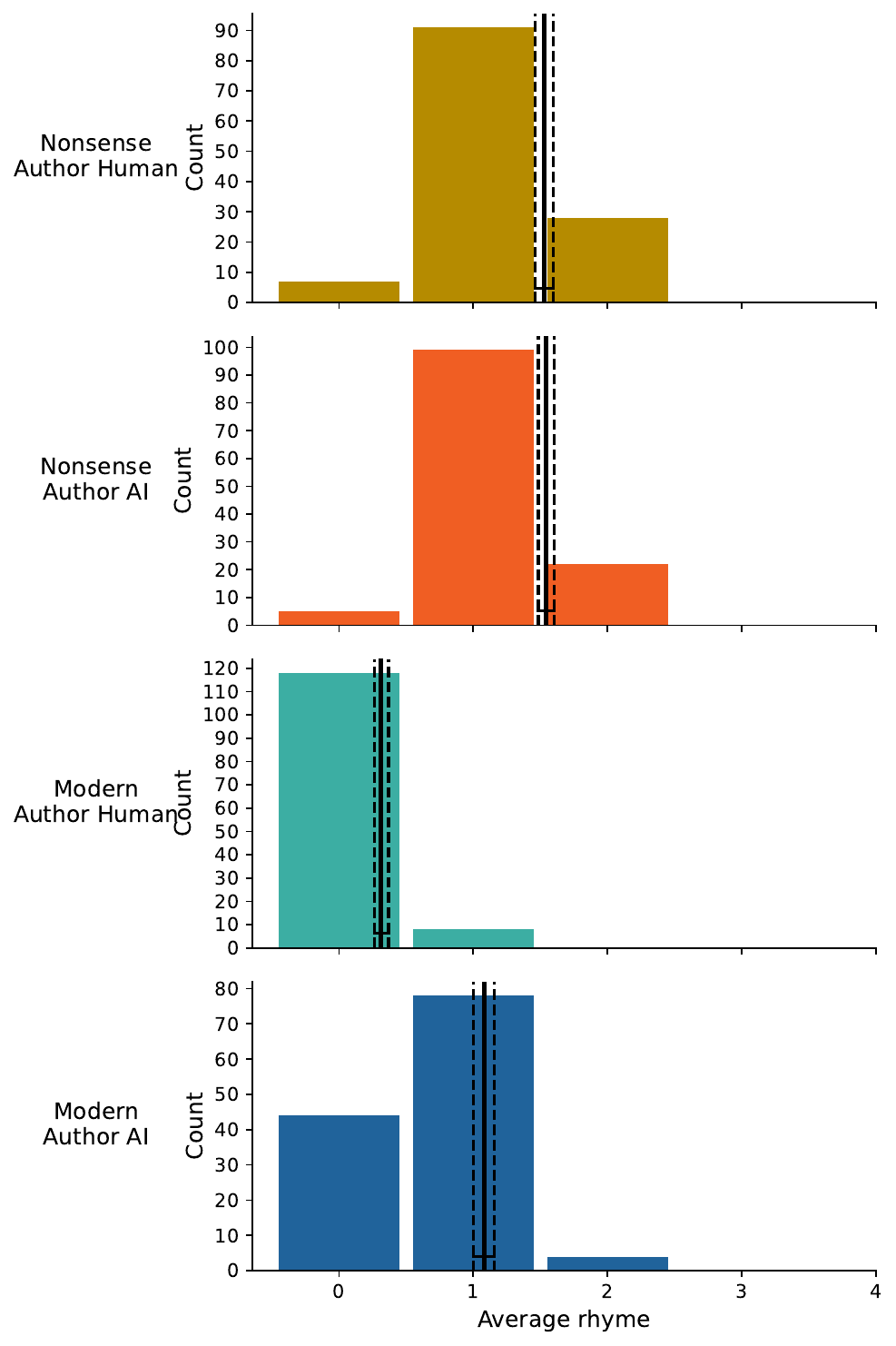}
       \caption{average rhyme according to poem authorship}
    \label{fig:rhyme_isauthorhuman_category}
\end{figure}

Lastly, we asked how well the poems rhyme. This aspect was more objective and less evaluative, as we aimed to examine whether AI can produce rhyming poetry in Czech. As shown in Figure~\ref{fig:rhyme_isauthorhuman_category}, when human-authored nonsense poems contained rhyme, AI was able to successfully mimic this pattern, with comparable ratings ($M = 1.54$, 95\% CI [1.48, 1.60] for AI; $M = 1.53$, 95\% CI [1.46, 1.59] for human). Additionally, while Czech modern poems written by humans were mostly non-rhyming ($M = 0.31$, 95\% CI [0.26, 0.37]), AI incorporated rhyme into modern poems as well, although to a lesser extent than in nonsense poems ($M = 1.08$, 95\% CI [1.00, 1.16]).


\section{Conclusions}
In this paper, we explored how native Czech speakers evaluate AI- and human-written poems. Our analysis leads to several conclusions.

First, large language models can convincingly mimic poetry even in a less-represented language, such as Czech, in terms of AI training data. Even though we used the most accessible method available to an ordinary user --- prompting GPT-4.5-Preview in chat mode without any hand-picking or post-editing of the outputs --- the resulting poems were already indistinguishable from human-written ones. This is an important finding, as the vast majority of studies on AI-generated poetry have focused on English. We believe that results based on English cannot be readily generalized to other languages, given English's unique position in AI training data. Research on languages that constitute less than 1\% of such data may offer a more accurate view of how AI models perform in other similarly underrepresented languages. So far, our experiment partly aligns with the findings of \citet{HITSUWARI2023107502}, who conducted an experiment on Japanese haiku and showed that AI-generated haiku with human interventions were rated stylistically higher than those written by humans. However, research on German poetry \citep{GunserEtAl2022} found that German AI poems were rated less favorably compared to human ones. In the future, our experiment should be replicated in additional low-resource languages to test the generalizability of these findings.

Second, LLMs appear to be slightly better at mimicking modern Czech poetry than nonsensical verse. This supports the idea that poetry is not a homogeneous genre. Although we observed a significant difference in participants' ability to recognize authorship between modern and nonsense poems, it is important to note that they were not particularly successful in identifying the nonsense poems either --- they were simply even less accurate with the modern ones.

Third, participants' experience with poetry did not influence their correctness in guessing authorship. We must note, however, that our sample did not include many professional poets or literary experts, which limits the conclusions that can be drawn. In our earlier pilot study (without evaluation scales), a large number of literature students similarly failed to distinguish AI poetry from human poetry, suggesting that expertise alone may not improve recognition accuracy. Nevertheless, this requires further investigation in a dedicated study.

Finally, the key finding of our study is that people rate poems lower when they believe they were generated by AI, even though there is either no actual difference in their evaluations of AI- and human-written poems, or AI-poetry is, in fact, evaluated more favorably. From our experiment, we cannot determine the causal direction --- whether people first assume that a poem is AI-generated and therefore judge it as less beautiful and playful, or whether they find it less beautiful and playful and thus infer it must have been written by AI. Whatever the direction, our results support the assumption proposed in the Introduction: the enjoyment of a poem and the knowledge or belief about its author are closely interconnected. Our findings are consistent with those of \citet{PorterMachery2024}, who likewise found that participants evaluated AI poetry more favorably, which may have led to misattributions of authorship.

While this finding could be seen as a pessimistic sign for the future of poetry and creative writing --- implying that people might stop writing poetry because AI can do it instead --- we argue that the emergence of AI should be viewed in the context of a broader, long-standing discussion about the role of the author. Is art only about the final product? Not necessarily. Across various artistic fields, including music and visual art, people tend to exhibit a hesitant, if not dismissive, attitude toward AI-generated works. 

If we return to Barthes's conception of authorship, an AI ``author'' turns out to be an almost textbook case of the writer who stepped aside for the reader. The author´s role, according to Barthes, is ``to combine the different kinds of writing, to oppose some by others, so as never to sustain himself by just one of them" \citep[][p.16]{Barthes1968}. Likewise, his description of the text as ``a space of many dimensions, in which are wedded and contested various kinds of writing, no one of which is original: [...] a tissue of citations, resulting from the thousand sources of culture" resembles an AI-generated output \citep[][p.16]{Barthes1968}. Yet the readers, whom Barthes places at the center of his account, demonstrate the persistent importance of authorship in our data. In other words, for ordinary readers, authorship itself remains part of what is being perceived: knowing or believing whether a work is human- or AI-written shapes its enjoyment. The author may be theoretically dead, but for the reader, it seems essential that they once actually lived.

\section{Competing interests}
No competing interest is declared.

\section{Author Contributions Statement}

\section{Funding}

Anna Marklová and Martina Vokáčová were supported by Primus Grant PRIMUS/25/SSH/010.
Jiří Milička was supported by the project Human-centred AI for a Sustainable and
Adaptive Society (reg. no.: CZ.02.01.01/00/23 025/0008691), co-funded by the European
Union.

\section{Acknowledgments}

\section{Declaration on using AI}
During the data analysis, the \emph{GPT-4}, \emph{GPT-4o}, and \emph{GPT-o1} models by \emph{OpenAI} and \emph{Claude 3.5 Sonnet} models by \emph{Anthropic} were consulted. However, all scripts underwent manual review and were, when necessary, corrected or further refined. By doing so, Jiří Milička assumes full responsibility for all the errors in these scripts. 

In the composition of this article, the \emph{GPT-5} was consulted for language styling and review. However, all ideas presented are original to the responsible authors, the generated text was further edited and underwent strict scrutiny.

\bibliographystyle{abbrvnat}
\bibliography{poetry}


\end{document}